\documentclass[preprint,12pt]{elsarticle}
\usepackage[T1]{fontenc}
\usepackage{lmodern}
\usepackage{amsmath}
\usepackage{graphicx}
\usepackage{booktabs}
\usepackage{tabularx}
\usepackage{siunitx}
\usepackage{xcolor}
\usepackage[colorlinks=true,linkcolor=blue!50!black,citecolor=blue!50!black,urlcolor=blue!50!black]{hyperref}
\biboptions{numbers,sort&compress}
\journal{arXiv}
\makeatletter
\def\ps@pprintTitle{\let\@oddhead\@empty\let\@evenhead\@empty
  \def\@oddfoot{\footnotesize\itshape Preprint \hfill \today}\let\@evenfoot\@oddfoot}
\makeatother

\begin{document}
\begin{frontmatter}

\title{When does deep reinforcement learning pay off in HVAC supervisory control?
Four small Modelica case studies trained directly on FMUs}

\author[1,2]{Bingnan Zhao\corref{cor1}}
\ead{bingnan.zhao@hdr.qut.edu.au}
\cortext[cor1]{Corresponding author.}
\author[1]{Harshala Gammulle}
\author[1]{Tharindu Fernando}

\affiliation[1]{organization={Signal Processing, Artificial Intelligence and Vision Technologies (SAIVT), Queensland University of Technology},
            addressline={2 George Street},
            city={Brisbane},
            postcode={4000},
            state={Queensland},
            country={Australia}}

\affiliation[2]{organization={International Laboratory for Air Quality and Health, Queensland University of Technology},
            addressline={2 George Street},
            city={Brisbane},
            postcode={4000},
            state={Queensland},
            country={Australia}}

\begin{abstract}
Deep reinforcement learning (DRL) has frequently been proposed for the supervisory control of
heating, ventilation and air-conditioning (HVAC) systems. The gains reported in the literature
are, however, difficult to interpret, because the baselines are often weak and the attainable
optimum is seldom known. This paper examines four small and fully reproducible case studies
based on an office building in Brisbane, Australia: pre-cooling under a time-of-use tariff;
chiller staging with condenser water reset; chilled water supply temperature reset with a wet
cooling coil; and chilled water storage under time-of-use prices and a demand charge. The
plants were modelled with the Modelica Buildings library, and in three of the cases a
soft actor-critic agent was trained directly on the exported functional mock-up unit (FMU).
Each agent was compared with tuned rule-based controllers and, where this was
practicable, with a near-optimal reference obtained either by hourly exhaustive search or by
dynamic programming with perfect foresight.

In the two cases without storage the agent reached the near-optimal reference, but the
margin above a well-chosen rule was small, namely 1.3\,\% for condenser water reset and
0.18\,\% for chilled water reset. In the latter case two common reset rules consumed more
energy than a fixed set point. With a chilled water storage tank the margin was considerably
larger. The agent outperformed both rules ($-7.8\,\%$ weekly cost, compared with $-3.9\,\%$ and
$-6.3\,\%$) and adopted midday charging during the solar off-peak period, yet it obtained
only about 45\,\% of the saving achieved by the perfect-foresight schedule and did little
to limit the demand peak. In the pre-cooling case a hand-tuned rule remained superior
($-12.0\,\%$ compared with $-5.6\,\%$). These results suggest that studies of DRL for HVAC
control should state the margin between the best rule and a near-optimal reference. Some
practical difficulties encountered in training on co-simulation FMUs are also reported.
The models, trained agents and scripts are openly available.
\end{abstract}

\begin{highlights}
\item Four open Modelica HVAC control problems with tuned rules and near-optimal references
\item Soft actor-critic agents were trained directly on co-simulation FMUs
\item Without storage, a well-chosen rule left a margin of only \mbox{0.2--1.3\,\%}
\item With chilled water storage, DRL outperformed both rules but not the optimum
\item Practical difficulties of training on co-simulation FMUs are reported
\end{highlights}


\end{frontmatter}

\section*{Nomenclature}
\begin{table}[!htbp]
\footnotesize
\begin{tabularx}{\textwidth}{@{}l>{\raggedright\arraybackslash}X@{\hspace{1.5em}}l>{\raggedright\arraybackslash}X@{}}
\multicolumn{4}{@{}l}{\textit{Abbreviations}}\\
AHU & air handling unit & HVAC & heating, ventilation and air-conditioning \\
CHW & chilled water & MPC & model predictive control \\
COP & coefficient of performance & RC & resistance--capacitance (model) \\
DRL & deep reinforcement learning & RL & reinforcement learning \\
FMI & Functional Mock-up Interface & SAC & soft actor-critic \\
FMU & functional mock-up unit & VAV & variable air volume \\[4pt]
\multicolumn{4}{@{}l}{\textit{Symbols}}\\
$a_t$ & action in hour $t$ & $r_t$ & reward in hour $t$ \\
$c_t$ & cost incurred in hour $t$ (AUD or kWh) & $s$ & reward scaling constant \\
$c_t^{\mathrm{ref}}$ & cost of the reference controller in hour $t$ & & \\
\end{tabularx}
\end{table}

\section{Introduction}
Supervisory control of HVAC systems determines set points and operating modes at intervals of
minutes to hours, while local feedback loops track these set points. Reinforcement learning
(RL) is attractive for this task because a controller can be learned from a simulator
without an explicit optimisation model, and a considerable body of work has explored the
idea~\citep{wei2017drl,zhang2019whole,wang2020rl,yu2021review}. Two features of this
literature make the reported results difficult to compare. First, the baseline is often a
fixed set point rather than the kind of rule that an experienced engineer would apply, such
as wet-bulb tracking of the condenser water temperature, outdoor-air reset, or the sequences
of ASHRAE Guideline~36~\citep{ashrae2021g36}. Second, the optimum is rarely known, so it is
not clear whether a reported saving is large or small relative to what was achievable.
Benchmarking frameworks such as BOPTEST~\citep{blum2021boptest} and its Gym
interface~\citep{arroyo2021gym} go a long way towards resolving the first difficulty for a
defined set of test cases.

The present paper approaches the problem from another direction. It considers four
deliberately small control problems, chosen so that a near-optimal reference could be
computed in most of them. Each problem uses an openly available Modelica model, at least one
tuned rule and, where practicable, an hourly exhaustive search or a dynamic programme with
perfect foresight. The aim is not to establish whether DRL is able to control HVAC plant,
which earlier work has already shown, but to examine under what conditions it is worth
applying. This is expressed by two quantities: the margin that a problem leaves above a good
rule, and the fraction of that margin which the agent obtains.

\section{Related work}
Reviews of RL for building control~\citep{wang2020rl,yu2021review} and of field
demonstrations~\citep{khabbazi2025field} discuss these questions at some length.
\citet{biemann2021eval} compared four actor-critic algorithms on a simulated data centre, and
several authors have compared RL with model predictive control (MPC) on a common
emulator~\citep{arroyo2022rlmpc,stoffel2022eval,wang2023rlmpc}, in which case MPC provides a
practical upper reference. Modelica models have been coupled to RL through FMUs, for instance
in ModelicaGym~\citep{lukianykhin2019modelicagym} and BOPTEST-Gym~\citep{arroyo2021gym}. At the
plant level, RL has been applied to cooling water systems~\citep{qiu2020cooling}, to chiller
plants~\citep{wang2025chiller}, and to active and passive thermal
storage~\citep{henze2003tes,liu2007tes}. For chilled water systems without storage,
\citet{braun1989chw} showed long ago that a simple control law can approach the optimum
closely. Chilled water reset based on outdoor temperature remains common practice and
continues to be evaluated in office buildings~\citep{lee2022reset}.

The contribution of the present paper is therefore not a new algorithm. It consists of a small set
of open, FMU-based problems in which the margin above a tuned rule is measured, together with
some practical observations on training RL agents directly on co-simulation FMUs.

\section{Methodology}
\subsection{Common framework}
All plants were modelled in Modelica using the Buildings library~\citep{wetter2014buildings},
version~12.0.0, and exported as FMI~2.0 co-simulation
units~\citep{blochwitz2011fmi} with the CVODE solver. Each FMU was wrapped as a Gymnasium
environment~\citep{towers2024gymnasium} and run through fmpy with a communication interval of
10~minutes; the supervisory decision interval was one hour. The agents were trained with soft
actor-critic (SAC)~\citep{haarnoja2018sac} as implemented in Stable-Baselines3~\citep{raffin2021sb3}.
In each case two random seeds were trained in parallel on a laptop CPU, which took between 9
and 25~minutes. The checkpoint that performed best on two validation periods was retained and
evaluated on a separate test period.

The weather data are those of the Brisbane Airport TMYx 2011--2025 file. The building is an
office occupied on weekdays from 07:00 to 18:00. Cases~2 and~3 minimise energy use. Cases~1
and~4 minimise electricity cost under the same time-of-use tariff, with an off-peak period from
11:00 to 16:00 at 0.08~AUD/kWh, a peak period from 16:00 to 21:00 at 0.48~AUD/kWh, and a
shoulder rate of 0.26~AUD/kWh at all other times.

\subsection{Reward with an exogenous reference}
The hourly cost is dominated by the building load, whereas the part that the controller can
influence is typically one to three orders of magnitude smaller. In every hour, therefore,
the cost of a reference controller operating under the same load and weather was subtracted
from the cost incurred by the agent:
\begin{equation}
r_t = -\bigl(c_t(a_t) - c_t^{\mathrm{ref}}\bigr)/s ,
\end{equation}
where $s$ is a scaling constant. Because the reference depends only on exogenous inputs, it
adds a policy-independent term to the return and leaves the optimal policy unchanged. In
Cases~2 and~4 the reference was evaluated with a quasi-steady Python approximation of the
plant, and in Case~3 with a second instance of the FMU running at the fixed set point. In
Case~2 this modification was necessary for learning to take place; it was retained in
Cases~3 and~4.

\subsection{Near-optimal references}
Where the plant carries little state from one hour to the next (Cases~2 and~3), an hourly
exhaustive search was performed: in every operating hour each candidate action was simulated
and the least costly one retained. In Case~3 the search was carried out on the FMU itself,
using a second instance that replayed the week up to the hour in question. For the storage
problem (Case~4) a dynamic programme over the state of charge was solved with perfect
foresight of load and weather. It was formulated on the quasi-steady approximation, and the
demand charge was handled by sweeping over a cap on the plant power. The resulting schedule
was then applied to the FMU in the same way as any other controller. Neither reference is a
strict bound: the hourly search is greedy, and the agent equalled or marginally improved on it,
while the dynamic programme optimises an approximation of the FMU rather than the FMU itself.
Both are accordingly described as near-optimal references.

\section{Case studies}
The four cases are summarised in Table~\ref{tab:summary}, and the results on the test periods
are given in Table~\ref{tab:results}.

\begin{table}[!htbp]
\centering\footnotesize
\caption{The four case studies.}
\label{tab:summary}
\begin{tabularx}{\textwidth}{@{}p{2.6cm}XXp{2.4cm}@{}}
\toprule
Case & Plant (Modelica) & Decision (hourly) & Coupling in time \\
\midrule
1 Pre-cooling & 100\,m$^2$ RC zone (air + mass), ideal chiller & Zone set point 20--30\,\textdegree C & Building mass \\
2 Condenser water & 2$\times$781\,kW screw chillers, variable-speed tower, 1.35\,MW load & Condenser water set point, staging & Weak \\
3 Chilled water & 781\,kW chiller, variable-flow CHW, wet AHU coil, VAV zone with moisture & CHW set point 5.5--11\,\textdegree C & Weak \\
4 Storage & Case~2 plant + 400\,m$^3$ stratified tank (3.4\,MWh) & Charge/discharge rate & Strong (tank, demand charge) \\
\bottomrule
\end{tabularx}
\end{table}

\begin{table}[!htbp]
\centering\footnotesize
\caption{Results on the test period (Case~1: three hottest consecutive weekdays; Cases~2--4: one summer week). Savings are relative to the
first row of each case; a negative saving means more energy than that row. The headroom is the difference between the best rule and the
near-optimal reference (hourly optimum or perfect-foresight schedule), expressed as a
percentage of the first row.}
\label{tab:results}
\begin{tabular}{@{}llrrl@{}}
\toprule
Case & Controller & Result & Saving & Remark \\
\midrule
1 & Fixed 24\,\textdegree C & 13.31 AUD & -- & \\
  & Expert pre-cooling rule & 11.70 AUD & 12.0\,\% & hand-tuned \\
  & DRL (SAC) & 12.56 AUD & 5.6\,\% & second seed 6.4\,\% \\
\midrule
2 & Fixed 29.4\,\textdegree C & \num{12225} kWh & -- & \\
  & Wet bulb + 4\,K & \num{12030} kWh & 1.6\,\% & \\
  & DRL (SAC) & \num{11865} kWh & 2.9\,\% & \\
  & Hourly optimum & \num{11871} kWh & 2.9\,\% & headroom 1.3\,\% \\
\midrule
3 & Fixed 6.67\,\textdegree C & \num{5606} kWh & -- & \\
  & Outdoor-air reset (tuned) & \num{5609} kWh & $-0.05$\,\% & \\
  & Pump trim \& respond & \num{5639} kWh & $-0.59$\,\% & \\
  & DRL (SAC) & \num{5596} kWh & 0.18\,\% & \\
  & Hourly optimum & \num{5596} kWh & 0.18\,\% & headroom 0.18\,\% \\
\midrule
4 & No storage & \num{3727} AUD & -- & \\
  & Time-of-use rule & \num{3583} AUD & 3.9\,\% & \\
  & Demand-limit rule (tuned) & \num{3492} AUD & 6.3\,\% & \\
  & DRL (SAC) & \num{3437} AUD & 7.8\,\% & second seed 4.0\,\% \\
  & Perfect foresight & \num{3077} AUD & 17.4\,\% & headroom 11.1\,\% \\
\bottomrule
\end{tabular}
\end{table}

\subsection{Case 1: pre-cooling under a time-of-use tariff}
A two-node RC model of an office zone, served by an idealised chiller, is exposed to three
consecutive hot weekdays. The agent sets the zone temperature set point, and the comfort
range during occupancy is 21--25\,\textdegree C. In this case the agent was trained on a
Python model with the same equations as the Modelica model; when its test episode was
replayed in Modelica, the zone temperature agreed to within 0.04\,K (root-mean-square).

\begin{figure}[!htbp]
\centering
\includegraphics[width=0.82\textwidth]{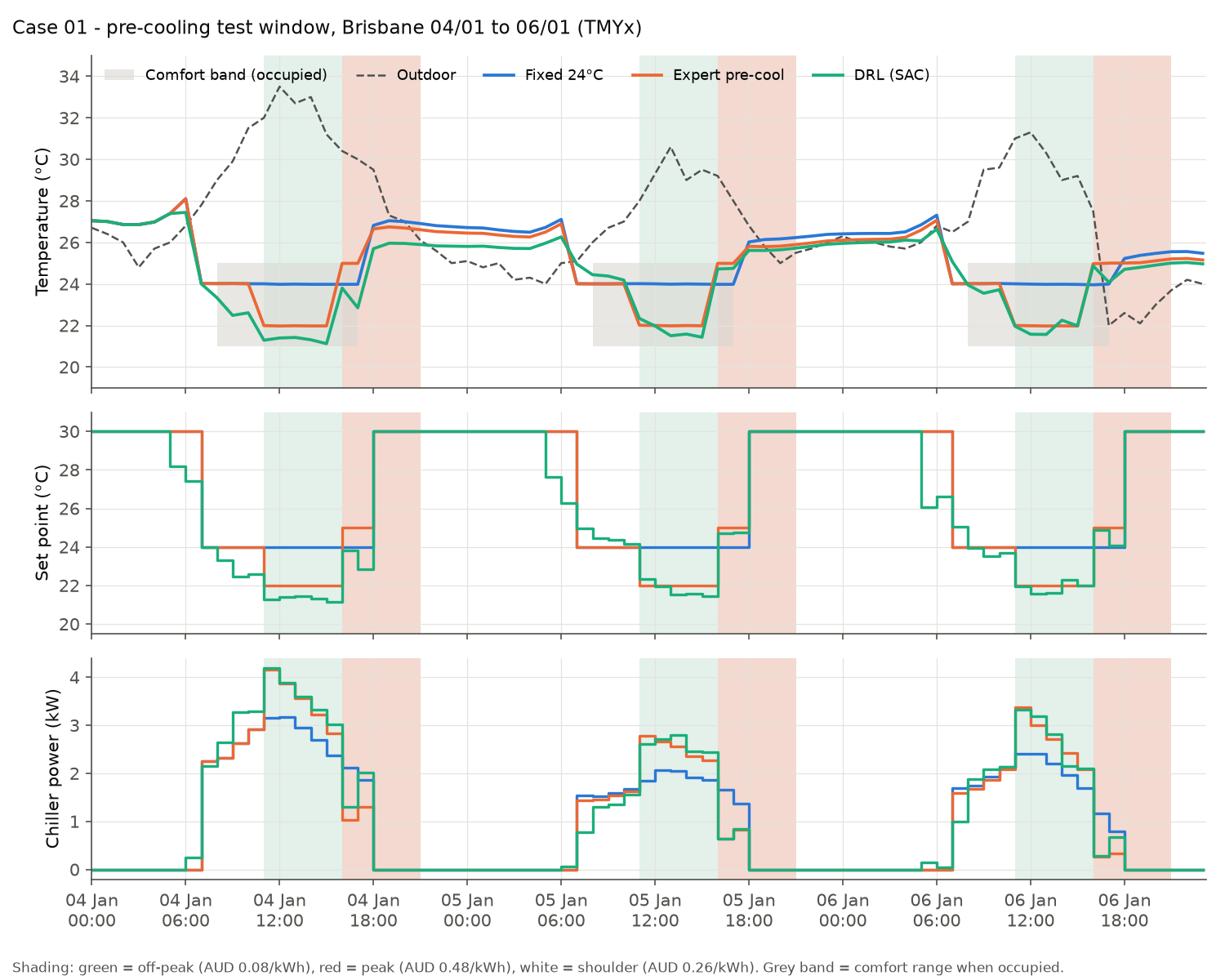}
\caption{Case~1: zone temperature, set point and chiller power over the three test days.
Green: off-peak; red: peak.}
\label{fig:c1}
\end{figure}

Without being instructed to do so, the agent learned to store cooling in the building mass
during the inexpensive midday period and to allow the zone temperature to rise during the
peak period (Fig.~\ref{fig:c1}). A simple expert rule, which holds 22\,\textdegree C during
the off-peak period and 25\,\textdegree C during the peak, follows the same strategy more
decisively and saves about twice as much. Plausible explanations are that the stored cooling
is only partly observable to the agent and that the price signal recurs only once a day. No
near-optimal reference was computed for this case.

\subsection{Case 2: chiller staging and condenser water reset}
\begin{figure}[!htbp]
\centering
\includegraphics[width=0.9\textwidth]{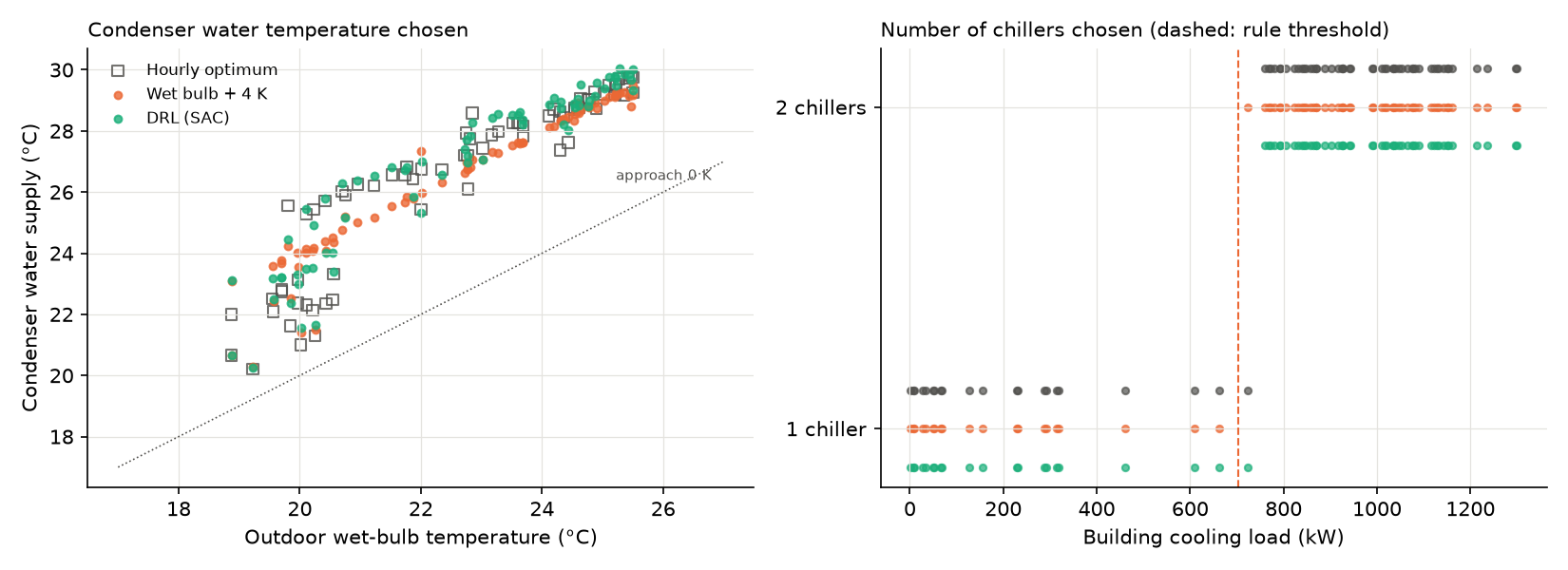}
\caption{Case~2: condenser water set point against wet bulb (left) and staging against load
(right) for the rule, the agent and the hourly optimum.}
\label{fig:c2}
\end{figure}

The agent selects the condenser water set point and the number of chillers in operation. It
arrived at essentially the rule used in practice, tracking the wet-bulb temperature with an
approach of about 4\,K and starting the second chiller at a load of about 720\,kW
(Fig.~\ref{fig:c2}), and its energy use was practically the same as that of the hourly
exhaustive search.
The entire margin above the tuned rule, however, was only 1.3\,\%. This is in keeping with
the finding of \citet{braun1989chw} that simple control laws come close to the optimum for
chilled water systems without storage.

\subsection{Case 3: chilled water supply temperature reset}
\begin{figure}[!htbp]
\centering
\includegraphics[width=0.9\textwidth]{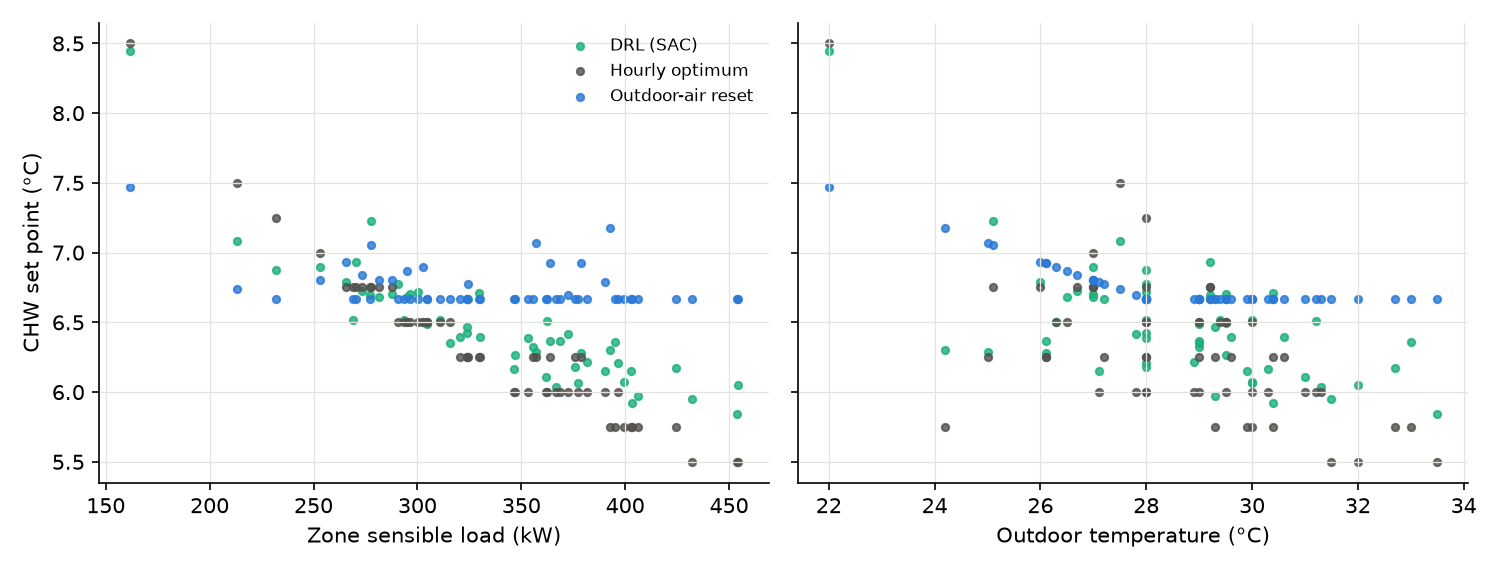}
\caption{Case~3: chilled water set point against zone load (left) and outdoor temperature
(right). The agent and the hourly optimum lower the set point as the load rises; the
outdoor-air reset does not.}
\label{fig:c3}
\end{figure}

Raising the chilled water temperature improves the COP of the chiller by roughly 3\,\% per kelvin.
In this plant, however, it also requires a larger water flow through the coil, and the pump
power rises with approximately the cube of the flow. Above about 8\,\textdegree C the pump
reaches full speed, the supply air temperature can no longer be held, and the supply fan,
whose power also varies roughly with the cube of the flow, accounts for the additional
losses. In consequence the best set point decreases as the load increases, from about
7.75\,\textdegree C below 250\,kW to about 5.6\,\textdegree C above 400\,kW. The agent learned
this relationship without any knowledge of the coil or pump characteristics
(Fig.~\ref{fig:c3}). Both reset rules, which raise the set point, used slightly more energy
than the fixed set point. The benefit of chilled water reset therefore appears to depend on
the size of the coil and on the share of pumping energy, and it would seem prudent to verify
it for a given plant rather than to assume it~\citep{lee2022reset}.

\subsection{Case 4: chilled water storage with a demand charge}
In this case the agent sets the rate at which the tank is charged or discharged in each
hour. A demand charge of 20~AUD/kW per month, pro-rated to one week, is levied on the highest
hourly plant power, and staging and condenser water reset follow the rule of Case~2.

\begin{figure}[!htbp]
\centering
\includegraphics[width=0.95\textwidth]{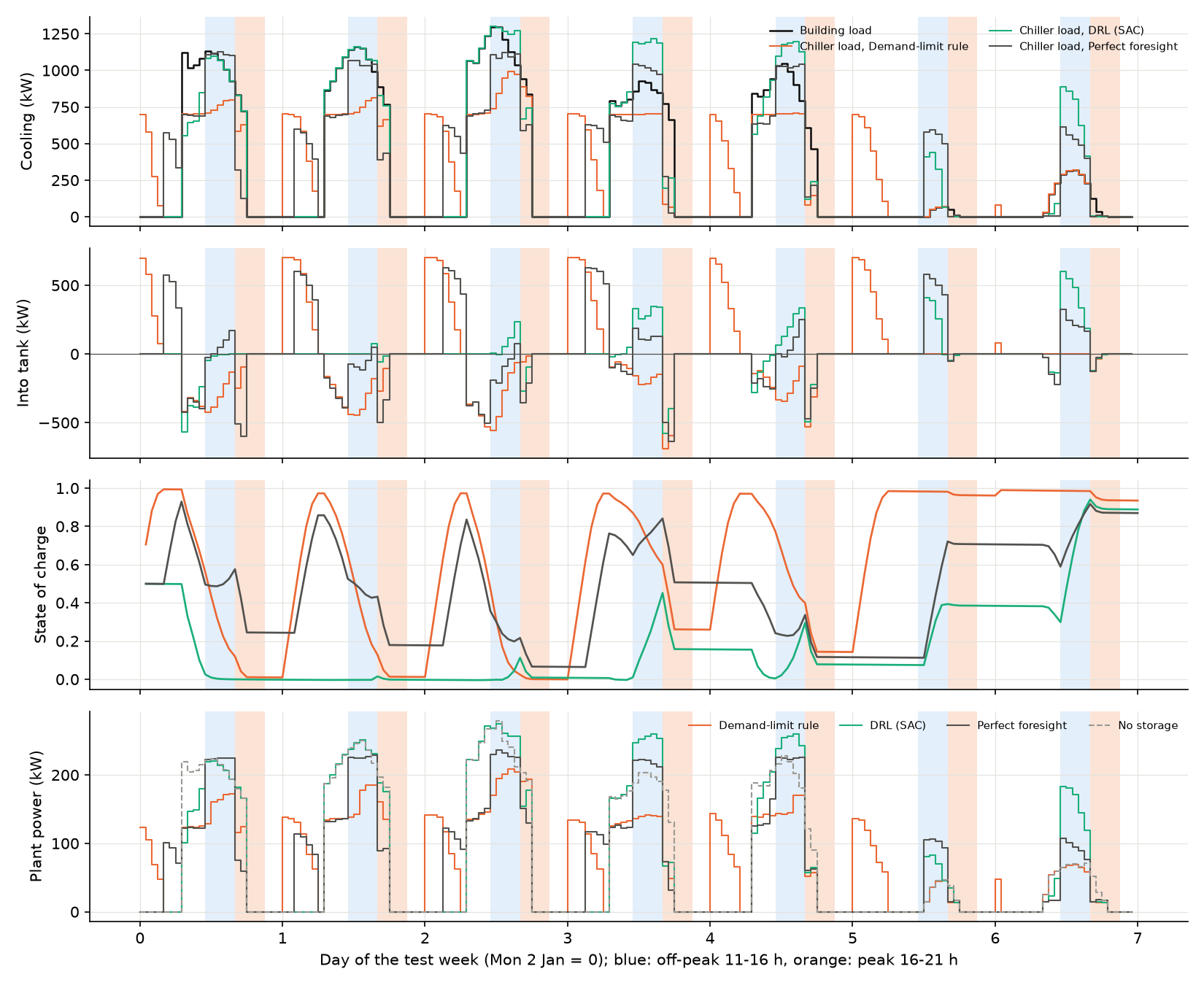}
\caption{Case~4: building and chiller load, tank heat flow, state of charge and plant power
over the test week for the demand-limit rule, the agent and the perfect-foresight schedule.}
\label{fig:c4}
\end{figure}

The agent achieved a lower cost than both rules (Fig.~\ref{fig:c4}), principally by charging
the tank during the midday off-peak period, when rooftop photovoltaic generation depresses the
price, rather than at night. Its weekly peak of 275\,kW was, however, scarcely lower than the
279\,kW of the plant without storage, whereas the demand-limiting rule reached 209\,kW and the
perfect-foresight schedule combined midday charging with a peak of 236\,kW. The demand charge
is determined by a single hour of the week, and in \num{100000} training steps the agent,
which acts at hourly intervals, did not learn to relate its earlier decisions to that hour.
This observation is consistent with earlier work on RL for thermal storage, which found that
the learned controller compared favourably with conventional strategies but did not reach the
performance of predictive optimal control~\citep{henze2003tes}, and that learning became slow
as the state and action spaces grew~\citep{liu2007tes}.

\section{Discussion}
\subsection{Margin above a good rule}
Cases~2 and~3 indicate that, for a supervisory problem with little storage and a sensible
existing rule, little remains to be gained. The agent reached the near-optimal reference, but
that reference was only 0.2--1.3\,\% better than the rule. An hourly exhaustive search of this
kind requires only minutes of computation, and it seems advisable to carry it out before a DRL
study of such a problem is undertaken. Case~3 also illustrates the opposite situation, in which
a widely used rule performed slightly worse than no reset at all; the reference made this
apparent.

\subsection{Problems with storage}
When decisions are coupled in time the margin was about ten times larger, and the agent found
a strategy that the rules did not use. Two features appear to have limited its performance:
cost terms determined by a single extreme hour, such as the demand charge, which a tuned rule
handled better; and storage that is only partly observed, as with the building mass in
Case~1, where a hand-tuned rule also performed better than the agent. In such problems the
fraction of the attainable saving obtained by the agent appears to be a more informative
measure than the saving relative to a baseline. In Case~4 the agent obtained about 45\,\% of
the saving of the perfect-foresight schedule, but only about 13\,\% of the margin above the
best rule (1.5 of 11.1 percentage points).

\subsection{Training on co-simulation FMUs}
Training directly on the FMU required only 0.6--5\,ms per simulated hour, so that no
surrogate model was needed. Several difficulties with the co-simulation FMUs used here
are worth recording. A change of input at a communication point did not trigger an event, so
that thresholds evaluated on internal variables, for example staging based on the chiller
load, retained stale values; in one instance the chillers remained off and the stagnant water
was heated beyond the validity range of the medium model. Discrete decisions were therefore
taken outside the FMU and passed to it as inputs, and the FMU was initialised with the inputs
of the first action. Furthermore, a state restored with \texttt{fmi2SetFMUstate} did not
respond correctly to subsequent input changes, so the exhaustive search replays the trajectory
instead. Any physically impossible source of free cooling in the model was quickly found and
exploited by the agent; reporting the unserved cooling and the maximum supply temperature for
every controller revealed such errors at once. Finally, the formulation of the action proved
important: an action for which zero means ``no intervention'' (the charging rate) was learned
much more readily than a set-point-like action (a target for the chiller load).

\subsection{Limitations}
Each case uses one building, one climate, one tariff and a single test period, and only two
training seeds. In Cases~2 and~3 the differences between the agent, the reference and the
best rule (0.2--1.3\,\%) are of the same order as the modelling error; in Case~3, for example,
the FMU and the continuous-time simulation of the same model differ by 0.6\,\%, and the chiller curves are extrapolated
above a leaving water temperature of 8.9\,\textdegree C. The conclusion that the margin is
small therefore appears robust, whereas the ranking of controllers with nearly equal energy
use should not be relied upon. The result of Case~3 also depends on design choices, notably
the coil size, the pump characteristic and the supply air set point of 13\,\textdegree C at
full cooling. The models are deliberately small, and some components are idealised, among
them the cooling tower in Case~3 and the weekly pro-rated demand charge in Case~4. The
forecasts available to the agents are perfect. The findings should accordingly be read as
statements about these particular plants rather than about DRL or reset strategies in general.

\section{Conclusion}
For four small Modelica plants, SAC agents learned sensible supervisory policies; in the two
cases without storage they matched the near-optimal reference. Whether the effort was
worthwhile depended more on the problem than on the algorithm. Without storage the margin
above a good rule was between 0.2 and 1.3\,\%; with storage it exceeded 10\,\%, of which the
agent obtained only a part. It is suggested that the margin be estimated before DRL is
applied, and that the fraction of it obtained be reported. The models, exported FMUs, trained
agents and scripts are available online~\citep{zhao2026hvacdrllab}. Should the FMUs not run in a
particular environment, they can be regenerated from the Modelica sources.

\section*{Code and data}
The Modelica models, exported FMUs, trained agents and scripts are available at\\
\url{https://github.com/Bingnan-Zhao/hvac-drl-lab}\\
and are archived at Zenodo, doi:\href{https://doi.org/10.5281/zenodo.23910085}{10.5281/zenodo.23910085} (MIT licence).

\bibliographystyle{unsrtnat}
\bibliography{refs}
\end{document}